\documentclass{article}
\usepackage[margin=1in]{geometry}
\usepackage{graphicx}
\usepackage{bm}
\usepackage{amsmath}
\usepackage{amsfonts}
\usepackage{amssymb}
\usepackage{mathtools}
\usepackage{float}
\usepackage[round]{natbib}
\usepackage{booktabs}
\usepackage{caption}                 
\begin{document}

\title{Physics-Informed Learning for Robust Acoustic Localization with Calibrated Uncertainty}
\author{
  Jennifer N. Kampe$^{1,2}$ \and
  Changwoo J. Lee$^{2}$ \and
  Xin Shen$^{2}$ \and
  Ari Lehtiö$^{3}$ \and
  Sandro von Brandenburg$^{3}$ \and
  Ossi Nokelainen$^{1,4}$ \and
  David B. Dunson$^{2}$ \and
  Otso Ovaskainen$^{1}$
}
\date{}
\maketitle

\begin{center}
\footnotesize
$^{1}$Department of Biological and Environmental Science, University of Jyväskylä, Jyväskylä, Finland\\
$^{2}$Department of Statistical Science, Duke University, Durham, NC, USA\\
$^{3}$Digital Services, University of Jyväskylä, Jyväskylä, Finland\\
$^{4}$Open Science Centre, University of Jyväskylä, Jyväskylä, Finland
\normalsize
\end{center}

\begin{abstract}
Recent advances in Passive Acoustic Monitoring (PAM) offer an opportunity to obtain ecological spatial point-process data at unprecedented scale. However, realizing this opportunity necessitates the development of accurate and scalable localization methods. In real-world outdoor soundscapes, however, the assumptions underlying classical localization methods such as hyperbolic and score-based localization are routinely violated by multipath dominance, near-field effects, and complex propagation. Under these conditions, classical localization methods become brittle, with extreme errors possible even in small detection arrays. Rather than statistically replacing the underlying physics, we propose a method to refine it and increase robustness outside of ideal operating conditions: a learned model operating on physics-informed acoustic features corrects a fast hyperbolic solver where it produces implausible solutions, substantially reducing catastrophic worst-case errors while matching its median accuracy on field data. We further provide calibrated, geometry-aware uncertainty estimates suitable for propagation into downstream spatial models. Evaluating on distributed microphone arrays in real and simulated outdoor environments, we demonstrate that the proposed method yields robust, uncertainty-aware localization, providing a step toward scalable automated wildlife monitoring in complex acoustic environments.
    
\end{abstract}

\section{Background}

Passive acoustic monitoring (PAM) with networks of autonomous recording units (ARUs) has become a widely used and minimally invasive means of monitoring vocal wildlife. Such devices are commonly used to characterize the distribution and abundance of the marine mammals, birds, bats, frogs, large terrestrial mammals, and certain insects at the level of the recording site through the classification of recorded vocalizations \citep{gibb2019emerging, sugai2019terrestrial}. A specialized and increasingly common application is acoustic localization, in which a vocalizing animal's position is estimated from the arrival-time delays of its calls across an array of time-synchronized receivers \citep{Rhinehart2020}. 

While the physics of sound source localization is well understood, real-world
localization tasks routinely violate the simplifying assumptions required by classical
methods. Classical localization relies on the Time Difference of Arrival (TDOA) across multiple synchronized receivers to locate a sound source. These delays are typically estimated by generalized cross-correlation with phase transform (GCC-PHAT), which cross-correlates a pair of receiver signals and takes the lag of the dominant peak as the estimated arrival-time difference---a step reliable in anechoic conditions but fragile in reverberant ones, where a reflected-path peak may exceed the direct path and yield a grossly incorrect delay \citep{cobos2020frequency}. A fixed arrival-time difference between one receiver pair corresponds to a fixed difference in distance to the two receivers, constraining the source to a hyperbola with the pair as foci. Under ideal conditions, the true location is recovered by intersecting the hyperbolae from all receiver pairs---an operation taking microseconds to milliseconds. With few microphone pairs and little measurement redundancy, a single grossly mis-selected peak can dominate the fit and displace the estimated location by large distances, leaving the method prone to occasional gross errors in cluttered, reverberant outdoor arrays.

An alternative family of grid- or score-based methods (e.g., steered-response power) instead searches over candidate locations, scoring each by how consistently the received signals align under its implied delays. See \cite{Grinstein2024} for a recent comprehensive review of score-based methods. In general, score-based methods are more robust to noise but substantially more expensive, since cost scales with the search grid rather than admitting a closed-form solve. They remain subject, however, to a further set of challenges that degrade both families.
 
 Key challenges to the above classical localization methods include multipath dominance \citep{kaneko2022large}, asymmetric
attenuation, and environmental model misspecification arising from unobserved or weakly
parameterized propagation conditions (e.g., temperature-dependent sound speed and habitat-dependent attenuation) \citep{LELLOUCH2025}. In addition, contamination by non-acoustic
interference---such as mechanical vibration, contact, rubbing, impulsive events, or
localized noise near individual sensors---causes hard failures in classical methods,
since these signals bypass or violate the assumed acoustic propagation model rather than
merely increasing measurement noise. Even under a correctly specified propagation model
and in the absence of such interference, limitations imposed by experimental design
(e.g., array geometry, sample rate, and bandwidth) can produce spatial aliasing, in which
distinct source locations yield indistinguishable measurements \citep{vantrees2002optimum}. In these regimes the
information required for localization is distributed across many weak, heterogeneous cues,
interspersed with misleading or inconsistent evidence. From a Bayesian perspective, these
effects correspond to likelihood misspecification and intrinsic non-identifiability,
yielding multimodal or biased posteriors that cannot be resolved through stronger priors
alone.

Physics-informed machine learning has shown strong potential to improve estimation in complex real-world systems that are governed by clear physical laws but whose parameters
are under-characterized or misspecified, for example in learning partial differential
equations \citep{xu2025partial} and spatial modeling of heavy-metal concentrations
\citep{LIU2023110863}. 

Tabular data, though ubiquitous in practice, is a challenging regime for modern deep
learning. Methods such as CNNs and transformers thrive where they can exploit strong structure---spatial in images, sequential in text---but tabular data offers little such
structure to inform an inductive bias: column order is arbitrary, and features are
heterogeneous and conceptually unrelated. As a result, flexible deep models often underperform simple gradient boosting. Gradient-boosted decision trees (XGBoost, LightGBM, CatBoost) fit a function by sequentially adding small trees, each correcting the residuals of the last, and are a strong default on tabular data. They are, however, greedy function approximators: they extrapolate poorly beyond the training distribution and provide no native uncertainty quantification --- limitations that matter for our setting, where calibrated uncertainty is a requirement.

Prior-Data Fitted Networks (PFNs) \citep{muller2021} provide a gradient-free alternative means of approximating a posterior predictive distribution (PPD). Whereas a standard transformer learns a single function mapping inputs to outputs, a PFN learns, across many synthetic datasets, to approximate the PPD under a
specified prior over data-generating mechanisms. Unlike standard deep learning on tabular data, a PFN does not train on the target dataset: inductive bias is supplied entirely by the pretraining prior, providing excellent performance in the low-structure tabular regime.
TabPFN \citep{hollmann2023, pfefferle2025} extends the method to tabular
classification and regression, pretrained on large collections of synthetic datasets
drawn from a random-function prior based on Structural Causal Models and Bayesian Neural
Networks, from which it acquires a broad, reusable prior over tabular problems.

At inference, the pretrained model is applied in a single forward pass to a set of
training examples (features paired with responses) together with the test features,
approximating the PPD for the test response given those training examples. Rows are treated as tokens, and \textit{attention} learns
which training examples are relevant to a given test point and which features co-vary
across rows. As a result attention is pairwise and therefore
scales quadratically in the number of rows.

\section{Related work}
Acoustic localization in bioacoustics is dominated by the hyperbolic (TDOA) approach: in a comprehensive review of the field, \citet{Rhinehart2020} found that 69 of 86 surveyed studies reporting a position-estimation algorithm used a method in this family. Hyperbolic localization estimates a source position from inter-receiver arrival-time differences across a distributed, wide-baseline array, and is the paradigm we adopt. A distinct family of direction-of-arrival (DOA) methods instead uses compact, closely spaced arrays to estimate a source bearing rather than a position, requiring multiple nodes or additional range information to localize; these address a different deployment regime and are outside our scope \citep{Rhinehart2020}.

More recently, fully automated pipelines have been developed to localize terrestrial wildlife at scale \citep{freeland2026fully}, combining automated detection and cross-receiver matching with explicit error-rejection stages that discard unreliable TDOA estimates. Our approach targets the same failure modes but, rather than discarding flagged detections, substitutes a learned estimate from a model over physics-informed features, paired with calibrated, geometry-aware uncertainty.

Conformal prediction has recently been used to attach distribution-free intervals to acoustic estimates: \citet{khurjekar2023uncertainty} apply it to deep-learning DOA estimation, \citet{khurjekar2024distribution} extend it to both DOA and source localization under noise, interference, and sensor-position uncertainty, and \citet{rozenfeld2025conformal} calibrate manifold-based localization with Gaussian processes.

\section{Method}

We propose a tabular PFN-based approach to acoustic localization, motivated by the strong empirical performance of PFNs across diverse tabular domains. Rather than replacing the underlying physics, we assume the relevant physical structure can be captured through carefully designed features, and use data-driven inference to learn how to combine weak cues, resolve ambiguities, recognize interference-induced failures, and compensate for forms of model misspecification that are difficult to express analytically. In this way, the burden shifts from explicitly modeling every failure mode to learning how the reliability of different physical cues varies across operating regimes. We use a geometry-aware correction that scales the conformal interval per detection by the geometric dilution of precision (GDOP), widening intervals where the array geometry is ill-conditioned. Our method leverages recent advances in tabular machine learning to improve upon classical localization while providing geometrically interpretable uncertainty quantification. 

Consider an acoustic array of 
$N$ synchronized receivers such that each target acoustic source produces up to 
$N$ detections, each recorded as a waveform. Rather than operating on the raw waveforms directly, we summarize each acoustic event as a row of physics-based features designed to expose physically meaningful cues to the model. These fall into several groups: \emph{timing features}, the inter-receiver time delays that are the core localization cue; \emph{uncertainty features}, which flag when those timing estimates are ambiguous or corrupted by multipath reflections; \emph{propagation features}, capturing how the environment attenuates and colors the sound; and \emph{array-geometry} and \emph{meteorological} covariates. 

We build three components on top of a hyperbolic solver and associated physics-derived features, each targeting a distinct part of the problem: a learned point estimate that refines the hyperbolic solver, a gate that substitutes this estimate where physical-plausibility checks flag the hyperbolic solution as unreliable, and a calibrated, geometry-aware uncertainty estimate that scales prediction intervals per detection by the geometric dilution of precision (GDOP), widening them where the array geometry is ill-conditioned.

\subsection{Training data simulation}
To generate labeled training data, we utilize a state-of-the art physics-based simulator \citep{shen2026forestirphysicsinformedforestsound}. \textbf{ForestIR} simulates the propagation of a sound source (e.g., a bird vocalization) to each receiver in a bioacoustic remote sensing array under specified forest and environmental conditions and array geometry. For a given source position, we model the principal propagation paths---the direct path, the ground reflection, and scattering off trunks and foliage---each delayed and attenuated according to physics. Together
these define an impulse response, i.e., the transformation applied to any sound traveling from that position to that receiver. Convolving a recorded \textit{dry} call with this impulse response yields the signal the receiver would have observed. 

For each site, synthetic source positions are drawn uniformly at random within the microphone array's convex hull, expanded outward by a prespecified proportion of the maximum pairwise microphone distance so that the model also sees positions just outside the array footprint rather than only interior ones. Each simulated recording pairs a bird call---drawn at random from a pool of real recordings of species whose ranges include the site---with a background-noise regime and weather condition sampled from ranges representative of the site. Noise is drawn from a small site-specific set including real environmental recordings, synthetic white, pink, and band-limited (300–3000 Hz, overlapping the range of most bird calls) noise. Temperature and relative humidity are sampled per batch from site-specific ranges spanning the conditions observed during the corresponding field campaign, since both affect the speed of sound and propagation loss. The scattering model is configured per site: each surveyed tree is represented as a single point scatterer at its trunk location, and scattering is disabled entirely for open, treeless sites. Additional information on the dry audio and other simulation parameters is available in \cite{shen2026forestirphysicsinformedforestsound}.

\subsection{Feature extraction}

\begin{table}[htp]
\centering
\caption{Current tabular feature representation used for acoustic localization. Each row corresponds to a single simulated or real acoustic experiment.}
\begin{tabular}{p{0.28\linewidth} p{0.65\linewidth}}
\hline
\textbf{Feature category} & \textbf{Description} \\
\hline
TDOA features &
Microphone--microphone time-difference-of-arrival estimates, including multiple GCC peak candidates rather than only the dominant peak. \\

Cross-correlation scores &
Pairwise cross-correlation and GCC-PHAT confidence measures, including band-limited variants to capture frequency-dependent timing behavior. \\

Uncertainty indicators &
Reliability measures derived from GCC structure (e.g., peak separability, confidence ratios, lag gaps) summarizing timing ambiguity and multipath effects. \\

Impulse response summaries &
Global and band-limited statistics of the impulse response such as energy decay, peak arrival times, and reverberation-related measures. \\

Spectral features &
Spectral centroid and related summaries capturing frequency-dependent attenuation, coloration, and distance effects. \\

Envelope-based delays &
Time shifts between amplitude envelopes across microphones, providing low-frequency delay cues complementary to phase-based TDOA features. \\

Geometry features &
Microphone array descriptors such as inter-microphone spacing and array scale, encoding spatial context independent of the signal. \\

Environmental parameters &
Recorded or simulated conditions including temperature and noise level, used to contextualize feature reliability. \\

Hyperbolic solver outputs &
Point estimate and fit diagnostics (e.g.\ residual, conditioning) from the closed-form
hyperbolic solver, used as input features for the point-estimate model only (excluded from
the uncertainty model). \\

Targets &
Source location coordinates $(x, y)$ (and optionally $z$) used as regression targets. \\
\hline
\end{tabular}
\label{tab:features}
\end{table}

Rather than expose the model to raw waveforms, which contain large amounts of information irrelevant to localization, we convert each acoustic event into a row of tabular features based on  quantities that classical theory identifies as informative for localization. These include microphone-pair time-difference-of-arrival estimates--- retaining multiple GCC peak candidates rather than only the dominant peak---and cross-correlation confidence scores that quantify how reliable each timing estimate is; reliability indicators derived from GCC structure that flag timing ambiguity and multipath; impulse-response and spectral summaries capturing reverberation, attenuation, and distance-dependent coloration; envelope-based delays complementary to phase-based TDOA; array-geometry and environmental descriptors that supply spatial and operating
context; and the outputs of the closed-form hyperbolic solver itself. Together these comprise 72 features per experiment, summarized in Table \ref{tab:features}.

\subsection{Hybrid point estimate}
For the point estimate, we use TabPFN to regress the source coordinates, fitting independent models for $x$ and $y$ coordinates. The  hyperbolic solver's own position estimate is included as an input feature, so the learned model refines the physics-based solution rather than localizing from scratch. To guard against the cases where the learned model would degrade a reliable solution, we retain the solver's estimate by default and substitute the TabPFN prediction only on detections where physical plausibility checks flag the solver as unreliable. 

Plausibility is assessed by a two-stage gate. In the first stage, the hyperbolic solver's estimate is compared against a bounding box containing microphone array, expanded outward by a tunable margin proportional to the array's spatial extent; any estimate falling outside this region is rejected unconditionally. The second stage applies only to estimates that survive the first stage, asking the question: \textit{are the observed loudness differences consistent with the proposed solution?} 

We assess this by computing an energy-consistency score as the RMS residual between each microphone pair's observed log-amplitude ratio and the ratio predicted from the candidate position under $1/r$ spherical spreading, whereby amplitude decays in proportion to distance from the source. We flag the estimate if this residual exceeds a threshold $\tau^*$. Because source loudness and microphone sensitivity cancel in the ratio, this check requires no absolute sound-level calibration. This is the amplitude-domain analogue of ratio-based energy localization, in which the ratio of received levels at a pair of sensors removes the dependence on source level and constrains the source location \citep{Cobos_survey, Meng2017energysurvey}; here we use it as a plausibility check on the solver rather than as a localizer. The threshold $\tau^*$ is selected on simulation data alone, as the most permissive value (by grid search) achieving at least a threefold improvement in 99th-percentile error for no more than a 20\% increase in median error relative to the unsubstituted solver, and
then applied unchanged to the field data with no tuning on field labels. 

A detection is substituted with the TabPFN prediction if either the geometric check is failed or the
energy-consistency residual exceeds $\tau^*$; otherwise the solver's estimate is kept. This gating bounds the catastrophic tail while preserving the solver's accuracy on the majority of detections where it is already trustworthy.

\subsection{Uncertainty quantification}
Finally, we provide uncertainty quantification through geometry-aware conformal calibration. While TabPFN provides a native predictive distribution, this reflects \emph{model} uncertainty (i.e., uncertainty about the learned function given the data) which is not geometrically interpretable. We therefore quantify uncertainty separately, by fitting a second TabPFN model that predicts the localization error of the hybrid estimate from a feature set excluding the hyperbolic point estimate.


The error model outputs predicted quantiles of the localization error, which carry no finite-sample coverage guarantee, so we calibrate them by conformal prediction. We consider two ways of sizing the conformal correction. The \emph{fixed} correction applies a single constant adjustment to every detection (i.e., standard split conformal), which yields marginal coverage, such that the nominal rate holds on average, but may be too low in geometrically hard regions and too high in easy ones. The second approach utilizes the geometric dilution of precision (GDOP), a standard measure of how much array geometry amplifies timing error into position error at a given location. The \emph{GDOP-scaled} correction is a normalized (locally weighted) variant, scaling the adjustment per detection by GDOP so that intervals widen where the array geometry is intrinsically ill-conditioned. In both cases calibration is performed with positions held out of the calibration set (leave-one-position-out), so that reported coverage reflects generalization to genuinely unseen locations rather than to positions the model has already observed.

Conformal prediction requires that calibration and test prediction errors be exchangeable. In our setting exchangeability holds at the level of source positions but not individual detections: detections from a common position share an acoustic signature and are not interchangeable with positions elsewhere. We therefore take the position, rather than the detection, as the exchangeable unit. Coverage is correspondingly conditional on the held-out position being exchangeable with the
calibration positions, and on the multi-species playback stimuli being exchangeable with the vocalizations encountered at deployment.

\section{Simulation and Field Experiments}

\subsection{Simulated and field data}

To evaluate the proposed method, we use a combination of simulated and real field playback experiments. We study two experimental sites: a
\emph{frozen-lake} site (low noise, no trees), which also has real playback recordings from
the Konnevesi field experiment at known source positions, and a \emph{forest} site
(heterogeneous, noisy terrain), for which only simulation
is currently available. For each site we generate simulated multichannel acoustic experiments using the ForestIR package, placing a
bird-vocalization source at random 2D locations relative to the fixed microphone array and
rendering each recording with the physics-based simulator. 

The frozen-lake field experiment was conducted at the Konnevesi Research
Station on February 24, 2026. The deployment used a three-microphone array
in an approximately equilateral geometry with approximately 50~m
inter-microphone spacing, with microphones and speakers at 1.2~m height.
Being open and treeless, its simulation disables trunk scattering. Simulated weather is sampled from the range
recorded during the experiment ($-19$ to $-6\,^\circ\mathrm{C}$, 70--95\% RH). Each
simulated recording draws one noise regime at random from a site-specific
set: the site's own recorded background, synthetic white, or pink noise,
each at a fixed noise level. This site
additionally has real playback recordings: across four settings, two played
a 16-species European stimulus set and two a 30-species North American set,
from six known speaker positions, including one with a known hardware
failure. Playback streams were synchronized via PPS timing to approximately
1~cm in propagation distance and segmented into single-signal clips of
2.18--11.00~s (mean 5.07~s).

The forest site experiment was conducted in the Arabiankorpi protected
old-growth forest on June 18, 2026. This deployment used a six-microphone
array with maximum mic-to-mic distance approximately 100~m. Trunk
scattering in the associated simulations is parameterized by tree-trunk
positions mapped from satellite imagery \citep{NLS_orthophotos}. Simulated weather is
sampled from the range recorded during the experiment ($10.4$ to
$14.3\,^\circ\mathrm{C}$, 61--79\% RH). Each simulated recording draws one noise
regime at random from a site-specific set: a generic forest background,
synthetic white, pink, or band-limited noise.

Because TabPFN conditions on an in-context training set rather than
training on all available data, we use an approximately 90/10 train--test split with the
training set capped at $1{,}000$ examples.
Using this budget, we evaluate three protocols that differ in what real data, if any,
contributes to training. Under the \emph{simulation-only} protocol, both training and
evaluation use simulated data, establishing an in-distribution ceiling under fully
controlled conditions. Under the \emph{field-evaluation} protocol, the model is trained on
simulation alone and evaluated on the real field recordings, with every field position withheld entirely from training. One position's recordings were affected by a known hardware failure and are excluded from the per-detection error summaries below, though retained in the full analysis. Under the \emph{leave-one-position-out} protocol, we
rotate over the field positions, each time withholding one position from training while
adding a fraction of the remaining positions' recordings to the simulated training budget,
and pool the held-out predictions across rotations.

\subsection{Frozen-lake site: out-of-sample localization accuracy}

Table~\ref{tab:frozen_lake} reports localization error on the frozen-lake site, under both simulation and real playback recordings. In both settings the model is trained on simulation only, and evaluated on positions held out of training, so the field numbers reflect genuine out-of-sample generalization to unseen locations. On the real recordings, the hybrid gate matches the hyperbolic solver's median error
(0.41~m) while improving the upper tail, reducing the 90th and 95th percentiles from 24.6~m and 42.1~m to 22.3~m and 35.3~m respectively. 

\begin{table}[htp]
  \centering
  \footnotesize
  \renewcommand{\arraystretch}{1.2}
  \caption{Localization error on the frozen-lake site (meters), as per-detection 2D
    Euclidean distance to the true position, on a 3-microphone array. Hardware-failure
    recordings are excluded. The \emph{gate} returns the hyperbolic solution by default
    and substitutes TabPFN only where physical checks flag the solver as unreliable.
    Bold indicates the best value in each column within a setting.}
  \label{tab:frozen_lake}
  \begin{tabular}{llrrrrrr}
    \toprule
    \textbf{Setting} & \textbf{Method} & $n$ & \textbf{mean} & \textbf{p50} & \textbf{p90} & \textbf{p95} & \textbf{p99} \\
    \midrule
    \multicolumn{8}{l}{\textit{Real recordings} (trained on simulation only)} \\
    & Hyperbolic solver & 449 & 5.13 & \textbf{0.41} & 24.6 & 42.1 & 43.7 \\
    & Gate (hybrid)     & 449 & \textbf{4.97} & \textbf{0.41} & \textbf{22.3} & \textbf{35.3} & 44.3 \\
    \addlinespace
    \multicolumn{8}{l}{\textit{Simulation}} \\
    & Hyperbolic solver & 112 & 0.17 & \textbf{0.04} & \textbf{0.21} & \textbf{0.36} & 3.88 \\
    & Gate (hybrid)     & 112 & 0.17 & 0.05 & 0.29 & 0.43 & \textbf{1.80} \\
    \bottomrule
  \end{tabular}
\end{table}

A notable gap separates the simulation and field results: simulation error is far
smaller than real-world error (e.g.\ median 0.04~m vs.\ 0.41~m for the solver). This is expected: the frozen lake is superficially clear, but
subject to near-surface micro-climatic effects that are difficult to model, so the simulator understates real timing noise. For this reason, it's particularly important that each field deployment should include a short calibration step---playback from known positions---because this real data anchors the model and its uncertainty estimates.

\subsection{Forest site: simulation results }

We next consider a second deployment site in mixed forest, instrumented with a denser
six-microphone array. This site is the intended target of our approach: unlike the
frozen lake, the forest introduces trunk scattering, so the
time-difference solver more frequently locks onto a reflected path rather than the
direct one, producing occasional gross localization errors. It is therefore where we expect the learned model's tail robustness to matter most.

The preliminary results presented here are simulation-only, using the array geometry and environmental conditions of a protected forest site; the collection and analysis of corresponding field data is left to future work. The two methods trade off along exactly the axis the hybrid is designed to exploit. The hyperbolic solver is sharper in the center, with a lower median (0.14~m vs.\ 0.38~m) and a better 90th percentile (3.27~m vs.\ 5.00~m). But its error tail explodes under forest multipath: the 95th and 99th percentiles reach 56.6~m and 71.4~m,against 9.26~m and 14.8~m for TabPFN. 

\begin{table}[htp]
  \centering
  \footnotesize
  \renewcommand{\arraystretch}{1.3}
  \caption{Localization error on the forest site (meters), simulation only, on the
    site's six-microphone array. Tail quantiles at $n=112$ rest on relatively few
    detections and are indicative rather than definitive. Bold indicates the better
    value in each column.}
  \label{tab:forest_sim}
  \begin{tabular}{lrrrrrr}
    \toprule
    \textbf{Method} & $n$ & \textbf{mean} & \textbf{p50} & \textbf{p90} & \textbf{p95} & \textbf{p99} \\
    \midrule
    Hyperbolic solver & 112 & 5.36 & \textbf{0.14} & \textbf{3.27} & 56.6 & 71.4 \\
    TabPFN            & 112 & \textbf{1.58} & 0.38 & 5.00 & \textbf{9.26} & \textbf{14.8} \\
    \bottomrule
  \end{tabular}
\end{table}

\subsection{Calibrated uncertainty for the hybrid estimator}

Table~\ref{tab:coverage} reports empirical coverage at three nominal levels under standard split conformal correction and GDOP-scaled conformal correction. Because calibration requires real playback, the results below refer to the frozen lake experiment, for which playback recordings are available. Standard (fixed) conformal calibration is well-calibrated at the 0.50 and 0.95 levels but under-covers at 0.80 (0.73 empirical against 0.80 nominal). Scaling the correction by geometric dilution of precision recovers most of this shortfall (0.78), while leaving the already-nominal levels essentially unchanged. Because GDOP is a purely geometric, training-free quantity, this improvement requires no additional data---it reallocates interval width toward the locations the array geometry makes intrinsically harder to resolve. 

\begin{table}[htp]
  \centering
  \footnotesize
  \renewcommand{\arraystretch}{1.1}
  \caption{Empirical coverage of prediction intervals at three nominal levels,
    on $n=449$ held-out field detections, with positions held out of conformal
    calibration. \emph{Fixed} is standard split conformal; \emph{GDOP-scaled} is
    normalized conformal with a geometric scale function.}
  \label{tab:coverage}
  \begin{tabular}{lcc}
    \toprule
    \textbf{Nominal} & \textbf{Fixed} & \textbf{GDOP-scaled} \\
    \midrule
    0.50 & 0.50 & 0.52 \\
    0.80 & 0.73 & \textbf{0.78} \\
    0.95 & 0.96 & 0.96 \\
    \bottomrule
  \end{tabular}
  \label{tab:coverage}
\end{table}

The residual under-coverage at the $0.80$ level is concentrated at a single position---a location which the array geometry already identifies as the most ill-conditioned---rather than being distributed evenly across the site. This is consistent with the geometric account above, but suggests that fully closing the gap may require playback data at additional positions.

We find that in addition to being well calibrated, the predicted error is discriminative: it ranks high- from low-error detections substantially better than the purely geometric baseline (Spearman $\rho \approx 0.53$ for the learned model, versus $\rho \approx 0.22$ for GDOP alone), so the intervals are both calibrated and adaptive to which detections are trustworthy.

\section{Discussion}
The proposed method uses a highly efficient tabular representation of acoustic data that exploits decades of signal-processing insight. While neural networks can learn directly from structured data such as audio, converting raw recordings into physically meaningful summaries makes the well-established physics of acoustic propagation explicit rather than forcing the model to relearn it, at orders of magnitude less data and computational cost. Further, by extracting only those features relevant to localization, we avoid exposing the model to irrelevant melodic content which may harm generalization. Additionally, this formulation naturally supports hybridization with classical methods: hyperbolic solver outputs and geometric conditioning metrics enter simply as additional features, whose utility can be evaluated directly.

Our preliminary results suggest the method substantially reduces the catastrophic errors characteristic of hyperbolic localization failures while sacrificing little median accuracy, even under clean propagation conditions most favorable to the solver. This tail advantage appears larger still in the preliminary forest results, the regime the method is designed for, though field validation is the subject of ongoing work.

Together these results point toward a scalable deployment pipeline. Each site is characterized once, recording array geometry, approximate tree locations, and sine-sweep impulse-response measurements. This information is used to build a site-specific
simulation to which the model is fit; a short playback experiment from known positions
then calibrates the uncertainty. Because tree positions can be mapped remotely from satellite imagery rather than surveyed on foot, the structural characterization of a site requires no additional fieldwork; higher-fidelity inputs from aerial or terrestrial LiDAR are a natural extension where available. Since nearly all of this is one-time setup, requiring no sophisticated equipment beyond the monitoring array itself, the pipeline scales across many
sites. This scalability is what makes it feasible to localize detections at the volume and across the
sites needed to support spatial-ecological modeling---while the calibration step ensures
those detections carry the uncertainty such models require to be propagated.

\section{Acknowledgements}
This research was supported by NSF award \#2426762 and Academy of Finland award \#367674.

\bibliographystyle{plainnat}
\bibliography{mybib}

\end{document}